\documentclass[letterpaper, 10 pt, conference]{ieeeconf}

\IEEEoverridecommandlockouts                              

\usepackage[english]{babel}

\usepackage{amsmath}
\usepackage{graphicx}
\usepackage[colorlinks=true, allcolors=blue]{hyperref}

\usepackage{caption}
\usepackage{subcaption}

\usepackage{algorithm}
\usepackage{algpseudocode}

\usepackage{xcolor}
\usepackage{soul}

\usepackage{todonotes}
\presetkeys{todonotes}{inline,backgroundcolor=pink,caption={}}{}

\title{\LARGE \bf
CARPE-ID: Continuously Adaptable Re-identification for Personalized Robot Assistance
}

\author{Federico~Rollo$^{\dagger,\ddagger,\S}$,
        Andrea~Zunino$^{\dagger,\ddagger}$, \\
        Nikolaos~Tsagarakis$^{\ddagger}$,
        Enrico~Mingo~Hoffman$^{\mathparagraph}$, and
        Arash~Ajoudani$^{\ddagger}$
\thanks{$^{\dagger}$Intelligent and Autonomous Systems, Leonardo Labs, Genoa, Italy}
\thanks{\ \ \footnotesize e-mail: {\tt\footnotesize\{name.surname\}.ext@leonardo.com}}
\thanks{$^{\ddagger}$HHCM \& HRII, Istituto Italiano di Tecnologia, Genoa, Italy}
\thanks{\ \ \footnotesize e-mail: {\tt\footnotesize\{name.surname\}@iit.it}}
\thanks{$^{\S}$Industrial Innovation, DISI, Università di Trento, Trento, Italy}
\thanks{\ \ \footnotesize e-mail: {\tt\footnotesize\{name.surname\}@unitn.it}}
\thanks{$^{\mathparagraph}$Université de Lorraine, CNRS, Inria, LORIA, Villers-lès-Nancy, France}
\thanks{\ \ \footnotesize e-mail: {\tt\footnotesize enrico.mingo-hoffman@inria.fr}}
}

\begin{document}
\maketitle

\thispagestyle{empty}
\pagestyle{empty}

\begin{abstract}
In today's Human-Robot Interaction (HRI) scenarios, a prevailing tendency exists to assume that the robot shall cooperate with the closest individual or that the scene involves merely a singular human actor. However, in realistic scenarios, such as shop floor operations, such an assumption may not hold and personalized target recognition by the robot in crowded environments is required.  To fulfil this requirement, in this work, we propose a person re-identification module based on continual visual adaptation techniques that ensure the robot's seamless cooperation with the appropriate individual even subject to varying visual appearances or partial or complete occlusions. We test the framework singularly using recorded videos in a laboratory environment and an HRI scenario, i.e., a person-following task by a mobile robot. The targets are asked to change their appearance during tracking and to disappear from the camera field of view to test the challenging cases of occlusion and outfit variations. We compare our framework with one of the state-of-the-art Multi-Object Tracking (MOT) methods and the results show that the CARPE-ID can accurately track each selected target throughout the experiments in all the cases (except two limit cases). At the same time, the s-o-t-a MOT has a mean of 4 tracking errors for each video.






\end{abstract}

\section{Introduction}
    \label{sec:intro}


To accomplish Human-Robot Interaction (HRI) tasks in human-populated environments, robots need to recognize their human counterparts and tailor their actions accordingly. Such a personalization step is generally overlooked in traditional HRI settings, which assume that the robot collaborates with the nearest person and that there won't be other human interventions/distractions during the task execution. This assumption is not always realistic, especially when a task involves mobility and a larger workspace. For this reason, future collaborative robots must have the capacity to identify and recognise their human counterparts to deliver personalisable assistance.

Earlier works to address this focused on an offline human identification step followed by online tracking, e.g., for an omnidirectional mobile robot to follow a person~\cite{rollo2023followme}. In another work~\cite{ye2023robot}, the authors introduced a re-identification application in the presence of partial occlusions. Despite the progress made, these frameworks lack real-time adaptation to target appearance changes and cannot recover re-identification after total occlusions. In fact, they are relatively robust in tracking a target in the camera field of view (FOV), but when the target is lost for a moment, they often need a re-initialization of the tracking through human actions~\cite{javed2022visual}~\cite{luo2021multiple}. 


Given these limitations, our objective is to minimize human intervention in robot recognition and (re)identification for the human counterparts. Our first attempt to achieve this was presented  in~\cite{rollo2023followme}. The limitation of this approach was that, if a person changed his/her appearance after the calibration step (\textit{e.g.,} by changing the outfit), the re-identification module was not able to track the target person. To tackle this challenge, in this work, we provide a novel re-identification module which uses a deep learning approach based on feature extraction and a continual adaptation to be compliant with the target appearance changes. The robot constantly acquires images during the tracking and uses the new person's learned appearance to continuously update an ideal target representation which is used to re-identify the target when the tracking fails.

We present an adaptive person re-identification layer that takes advantage of a Multi-Object Tracking (MOT) algorithm\footnote{yolo\_tracking (MOT): \href{https://github.com/mikel-brostrom/yolo_tracking}{link}} based on the well-known YOLO framework \cite{redmon2016you} and the StrongSORT tracker \cite{du2023strongsort}, to supply a Re-identification system capable to handle the IDs' jumps that often occur in MOT\footnote{A detected object could switch the identification number (ID) due to partial and total occlusion or quick appearance changes.}.

The contributions of this work can be summarized in:
\begin{itemize}
    \item proposing a Continual Adaptation Personalized Re-identification framework\footnote{The authors will grant access to the source code based on the GPL 3.0 license, upon paper acceptance.}(CARPE-ID) to accomplish HRI tasks with a specific target;
    \item testing the tracking algorithm with a real Human-Robot collaborative scenario;
    \item evaluating limitations and possible solutions.
\end{itemize}
Videos of experiments can be found in the playlist at \href{https://youtube.com/playlist?list=PLdibjJfM06ztpaKqJhrkF3fJxqXhRwL00}{\textit{this link}}.

The rest of the paper is structured as follows. In Sect~\ref{sec:works} the state-of-the-art in Single and Multi-Object Tracking and HRI re-identification works and surveys are reviewed. 
Our approach is presented in Sect~\ref{sec:method}. Experiments and results are reported in Sect~\ref{sec:experiments} with a discussion on achievements, performances, and limitations. Finally, in Sect~\ref{sec:conclusion} the whole work is summarized and conclusive statements are presented.

\section{Related Works}
    \label{sec:works}

Many systems were proposed in the literature to cope with the object-tracking objectives. Multi-object tracking (MOT) and Single-object Tracking (SOT) are commonly used methods to track all objects or a single selected object in the image, respectively. MOTs algorithms generally perform both detection and tracking of multiple classes, while SOTs require only the initial guess (\textit{i.e.}, the object bounding box) to track the object of interest in that specific area, no matter which is its class type.
 
Based on a recent survey \cite{soleimanitaleb2022single}, the SOT techniques can be divided into four categories: \textit{(i)} feature-based, \textit{(ii)} segmentation-based, \textit{(iii)} estimation-based and \textit{(iv)} learning-based. They focused on learning-based approaches and then presented an overview of the datasets and the metrics used to evaluate the SOT algorithms. Similarly, the authors in \cite{zhang2021recent} analyzed the recent advancements in SOT focusing on correlation-based and deep learning-based algorithms while a broader analysis of Discriminative Correlation Filters and Siamese Network for Visual Object Tracking is presented in \cite{javed2022visual}. Instead, in the review presented in \cite{luo2021multiple}, the recent evolution of MOT systems and their applications are presented.

Considering the objectives of SOT and MOT algorithms, it is clear that they do not fulfil the requirements of personalisable HRI since they often lack robustness in the target re-identification after partial or total occlusions. For this reason, other approaches for person re-identification are proposed in the literature, especially for personalized robotics applications.

The most common approach for personalized re-identification for robotic assistance uses skeleton or face cues to track the target. For example, the authors in \cite{patruno2019people} proposed the use of soft biometric features extracted from depth and colour to build an informative descriptor of the target. They use 3D skeleton points extracted with AlphaPose~\cite{fang2022alphapose} to create a skeleton standard posture which is used to partition the target with a grid. For each grid cell, the mean colour is used for the re-identification, but it is limiting because, when people wear similar clothes, it could be prone to error. The authors in \cite{ye2023robot} developed a robot person-following robotic system which is robust to the partial occlusion often caused by the limited camera FOV. They used the target person's skeleton and a prior model to identify and localize it. The tracking and the identification are based on skeleton heuristics information such as joints, height, and bounding box position to identify the person. Even if this work is promising for a person-following scenario, a big assumption they make is that the person remains always inside the camera FOV. 

Regarding face recognition for person re-identification and tracking, the authors in \cite{liu2017online} train a metric model offline using labelled data, then use online face information to match the target and update the metric model. They propose the feature funnel model (FFM) to merge appearance information to the skeleton one. Also, in the work presented in \cite{wang2019real}, the authors use a face Re-id approach to enhance human-robot interaction. They pre-trained a feature extractor CNN that is used to re-identify target faces in an unsupervised manner. Both these works assume that the human collaborator always faces the robot though this is not always the case as we have seen in our experiments. 

The authors in \cite{koide2020monocular} use an RGB monocular approach. They first use OpenPose~\cite{cao2017realtime} for the skeleton extraction and then track the target using an unscented Kalman filter due to the information of the ground plane prior and the person height estimation. They combine convolutional channel features and online boosting to identify the person through deep features. The proposed method at the start needs a calibration process to learn the deep feature appearance that will be used to re-identify the target.
A distinct approach is proposed in \cite{cocsar2020human}, where the authors use a thermal camera to track and re-identify the target. They acquire a small dataset of thermal images to train a neural network and then use an entropy-based sampling to obtain a thermal dictionary for each person. A Support Vector Machine (SVM) classifier is trained using some geometric distribution obtained by the thermal dictionaries and then used to classify the people. The integration of a thermal camera is intriguing for solving standard camera problems. However, the need for extensive training and its inability to adapt to target changes restrict its applicability to our intentions.

As reported in the literature review, it becomes evident that all these works have certain limitations, which could hinder the development of a robust and personalized person-tracking system. Our approach solves online the problem of re-identification after partial and total occlusion and target appearance changes which, to the best of the authors' knowledge, is still an open problem for tracking systems in HRI applications.

\section{Methodology}
    \label{sec:method}
    
\begin{figure*}
    \centering
    \includegraphics[width=0.85\linewidth]{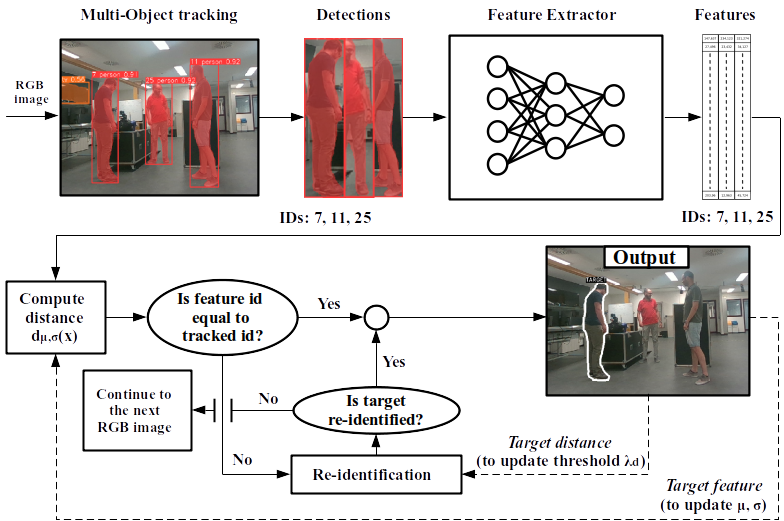}
    \caption{The figure shows the framework pipeline starting from the image input to the target re-identification output. The first module is MOT, where a neural network gives the first coarse tracking of objects in the image. 
    From the MOT module, we obtain the detections that are passed into a feature extractor outputting the output feature vectors $\boldsymbol{x_i}$ where $\boldsymbol{i}$ goes from zero to the number of the input detections. 
    The feature vectors $\boldsymbol{x_i}$ permits to compute the statistical distance $d_{\boldsymbol{\mu},\boldsymbol{\sigma}}(\mathbf{x})$ (see Eq.~\ref{eq:feat_dist}) that will be used by the re-identification module. 
    If the MOT can correctly track the target, we proceed directly to use its output, otherwise, if no target is found by the MOT, our re-identification module searches for a correspondence between the ideal target and the ones in the image. If the target keeps the same ID or if it is re-identified, the target statistical distance $d_{\boldsymbol{\mu},\boldsymbol{\sigma}}(\mathbf{x})$ and the target features $\boldsymbol{x_i}$ are used to update respectively the re-identifier threshold $\lambda_d$ and the ideal target representation $\boldsymbol{\mu}$ and $\boldsymbol{\sigma}$ (dashed lines). Otherwise, if no target is re-identified, the framework does not give any output and skips to the next RGB frame.}
    \label{fig:pipeline}
\end{figure*}

Our person-tracking method is represented in the pipeline in Fig.~\ref{fig:pipeline} and a pseudocode implementation is proposed in Algorithm~\ref{alg:carpe-id}. The tracking starts when an ID is selected by the user, meaning that before the ID selection, we run the MOT inference and store detected person appearances into a database to let the user choose the person to track.
After the ID selection, the initial tracking step is to acquire an RGB image and pass it into the MOT algorithm which provides an instance segmentation and a first object tracking, assigning different IDs to each detection. For each detection, the features representing their appearances are extracted with a deep neural network trained for person recognition (see details in Sect.~\ref{sec:experiments}) and then used to track or re-identify the target. 
These steps are repeated during the application which is conceptually divided into two steps that run sequentially for each frame: (i) target re-identification and (ii) target ideal representation update.
\begin{algorithm}
\caption{CARPE-ID framework}\label{alg:carpe-id}
\begin{algorithmic}[1]
\Statex $\textbf{Input:}\ tracked\_id$ 
\State $\boldsymbol{\mu}, \boldsymbol{\sigma}, \mu_d, \sigma_d, \lambda_d \gets initialize\_DEMA\_variables()$
\While{True}
\State $\boldsymbol{I_{rgb}} \gets camera.get\_rgb\_image()$
\State $detections \gets mot\_network.inference(\boldsymbol{I_{rgb}})$
\State $features \gets reid\_network.inference(detections)$
\State $min\_dist \gets MAX\_FLOAT\_NUMBER$
\State $tracked\_feat \gets Null$
\Statex {\textbf{\textit{\(\triangleright\) Target Re-identification}}}
\For{$feat \in features$}
    \If{$tracked\_id = feat.id$}
        \State $tracked\_feat \gets feat$
        \State \textbf{break}
    \Else 
        \State $d_{\boldsymbol{\mu},\boldsymbol{\sigma}} \gets get\_distance(feat,  \boldsymbol{\mu}, \boldsymbol{\sigma}), $
        \If{$d_{\boldsymbol{\mu},\boldsymbol{\sigma}} < min\_dist \And d_{\boldsymbol{\mu},\boldsymbol{\sigma}} < \lambda_d$}
            \State $min\_dist \gets d_{\boldsymbol{\mu},\boldsymbol{\sigma}}$
            \State $tracked\_feat \gets feat$
            \State $tracked\_id \gets feat.id$
        \EndIf
    \EndIf
\EndFor
\Statex {\(\triangleright\)\textit{\textbf{ Target Ideal Representation Update}}}
\If{$target\_feat \neq Null$} 
    \State $var \gets compute\_variance(\boldsymbol{\mu}, tracked\_feat)$
    \State $\boldsymbol{\mu} \gets DEMA(\boldsymbol{\mu}, tracked\_feat, \Delta_{f})$
    \State $\boldsymbol{\sigma} \gets DEMA(\boldsymbol{\sigma}, var, \Delta_{f})$
    \State $var_d \gets compute\_variance(\mu_d, min\_dist)$
    \State $\mu_d \gets DEMA(\mu_d, min\_dist, \Delta_{\lambda_d})$
    \State $\sigma_d \gets DEMA(\sigma_d, var_d, \Delta_{\lambda_d})$
    \State $\lambda_d \gets \mu_d + 2 \sigma_d$
\EndIf
\EndWhile
\end{algorithmic}
\end{algorithm}
\subsection{Target re-identification} \label{subsec:reid}
The features $\mathbf{x}$ extracted from the people detections are used to compute the statistical distance: 

\begin{equation} \label{eq:feat_dist}
     d_{\boldsymbol{\mu},\boldsymbol{\sigma}}(\mathbf{x}^*) = \sqrt{\frac{1}{D}\sum_{i=1}^D \left(\frac{\boldsymbol{x^*_i}-\boldsymbol{\mu}_i}{\boldsymbol{\sigma}_i}\right)^2}\text{,}
\end{equation}
where the subscript $\boldsymbol{i}$ represents the $i$-th index of the corresponding vector and $\boldsymbol{\mu}$ and $\boldsymbol{\sigma}$ correspond to the distribution representing the ideal target we want to track.
In the following step, we check if there exists a feature vector with an associated ID equal to the user-picked one; if this is the case then we directly output the detection paired with its ID. Otherwise, the re-identification module finds the feature with the smallest distance from the target ideal representation and checks if this distance is less than the adaptive threshold $\lambda_d$. If no person is re-identified among the detected ones then the framework continues to analyze the next RGB frame.
\subsection{Target ideal representation update} \label{subsec:mod_update}
Once the target is re-identified, we use its feature vector and corresponding statistical distance to continuously adapt the ideal target representation $\boldsymbol{\mu}$ and $\boldsymbol{\sigma}$ and the threshold $\lambda_d$, used in the re-identification step, which is computed as:
\begin{equation}
	\lambda_d = \mu_d + 2\ \sigma_d\text{,}
\end{equation}
where $\mu_d$ and $\sigma_d$ are the mean and the variance of the statistical distances of the target.
To update online the variables $\boldsymbol{\mu}$, $\boldsymbol{\sigma}$, $\mu_d$ and $\sigma_d$ we used a Damped version of the Exponential Moving Average $\boldsymbol{\chi}_{dema}$ that we will call DEMA now on. The DEMA formula is expressed in the following equation: 
\begin{equation} \label{eq:dema}
	\boldsymbol{\chi}_{dema}[k] =  \alpha_{damp}\ \boldsymbol{\psi} + (1 - \alpha_{damp})\ \boldsymbol{\chi}_{dema}[k-1]\text{,}
\end{equation}
where $\boldsymbol{\psi}$ is the new value, $[k]$ represents the discrete-time instant of the DEMA and $\alpha_{damp}$ is an adaptive weight which encodes the importance of the new values $\psi$ with respect to the previous DEMA value $\boldsymbol{\chi}_{dema}[k-1]$ and is computed as: 
\begin{equation}
	\alpha_{damp} = \frac{2}{N_{damp} + 1}\text{,}
\end{equation}
with $N_{damp} = N\cdot \Delta$, 
where $N$ is the number of the DEMA updates made till now and $\Delta$ is the damping factor that, as we will explain later, depends on what kind of information we are trying to update. It is a good practice to initialize $N=0$ to speed up the initial convergence and to set a fixed upper-bound $N_{MAX}$ otherwise as time passes the DEMA won't be affected by new values because as $N$ increases, $\alpha_{damp}$ tends to $0$.

\begin{figure}
    \centering
    \begin{subfigure}[b]{0.5\textwidth}
        \centering
        \includegraphics[width=\linewidth]{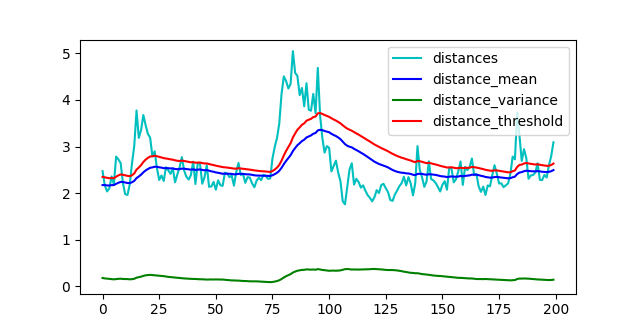}
        \caption{Representation of the effects of the EMA function.}
        \label{fig:no_damp}
    \end{subfigure}
    \vfill
    \begin{subfigure}[b]{0.5\textwidth}
        \centering
        \includegraphics[width=\linewidth]{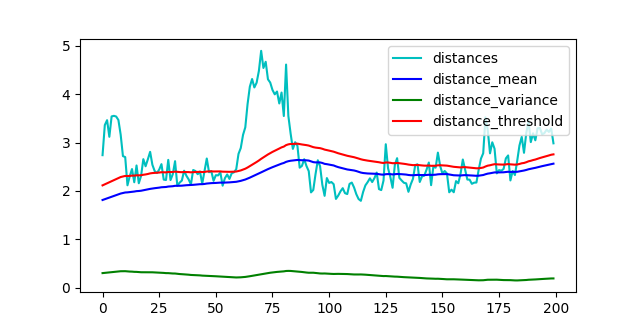}
        \caption{Representation of the effects of the DEMA function.}
        \label{fig:damp}
    \end{subfigure}
    \caption{Comparison between DEMA and EMA effects on threshold filtering and ideal target representation. In subfigure (a) the distances mean, variance, and threshold (in dark blue, green, and red) closely follow the behaviour of the distances plot (light blue line) with a small delay. In subfigure (b) the damping factor ensures that the mean, variance, and threshold do not follow the behaviour of the distance during big peaks and wait for the ideal representation to adapt to the new appearances of the target.}

    \label{fig:ema_comparison}
\end{figure}

There are two different damping factors $\Delta$ for the DEMA, one for the ideal target features represented by $\boldsymbol{\mu}$ and $\boldsymbol{\sigma}$ called $\Delta_{f}$, see~\eqref{eq:deltaf}, and one for the threshold $\lambda_d$ computed with $\mu_d$ and $\sigma_d$ that is called $\Delta_{\lambda_d}$, see~\eqref{eq:deltal}. In Fig.~\ref{fig:ema_comparison}, the EMA with and without the damping factors are compared. This image shows how the damping factor $\Delta_{\lambda_d}$ removes high-frequency noise from the threshold $\lambda_d$ (in red) and smooths its behaviour mitigating the EMA peaks that could lead to wrong re-identifications when the threshold becomes too high.

\begin{align}
    \Delta_{f} &= min(1,\ \frac{d_{\boldsymbol{\mu},\boldsymbol{\sigma}}}{2})\text{,} \label{eq:deltaf}
    \\
    \Delta_{\lambda_d} &= max(1,\ 2\frac{d_{\boldsymbol{\mu},\boldsymbol{\sigma}}} {\lambda_d})\text{.} \label{eq:deltal}
\end{align}

Substituting the damping factor $\Delta_{f}$ in~\eqref{eq:deltaf} in DEMA~\eqref{eq:dema} and $N_{damp}$, we are disfavouring that situations in which the distance in~\eqref{eq:feat_dist} from the current feature is small because in that case, the current feature is similar to the ideal one and this will overfit the ideal representation, especially in a situation in which the target is stationary in the same pose.
Instead, the damping factor $\Delta_{\lambda_d}$ in~\eqref{eq:deltal}, is used to avoid the excessive growth of the threshold when the ratio between the distance $d_{\boldsymbol{\mu},\boldsymbol{\sigma}}$ and the threshold $\lambda_d$ is too large, leading to wrong re-identification.

To improve the robustness of the algorithm we added a \emph{blacklist manager} that evaluates which IDs provided by the MOT are associated with distractors, to be able to discard in advance them during the re-identification step. When the target ID provided by the MOT does not change during the tracking, the manager adds the IDs associated with the other people present in the image (the distractors) to the blacklist. In this way, we nullified the few re-identification errors we could have in challenging situations.

\section{Experiments}
    \label{sec:experiments}
    
To validate our framework we performed two different experiments. We validate the framework per se using a fixed camera that acquires videos in a laboratory setup from different view perspectives (see sect.~\ref{subsec:individual}) and, to further validate the proposed method, we consider a second experiment that shows the use of the framework in a human-robot interaction application, \textit{i.e.} a robot person-following scenario (see sect.~\ref{subsec:whole}). Finally, in sect.~\ref{subsect:res_discussion}, we discuss the results obtained in the experiments.

The framework runs on a notebook with an \textit{Intel® Core™ i9-11950H} processor and an \textit{NVIDIA Geforce RTX 3080 Laptop} GPU. We used an Intel Realsense D455 camera for image acquisition and a Robotnik RB-Kairos+ 5e as the assistant robot. For the feature extraction network, we pre-trained an IBN-ResNet-50~\cite{pan2018two} on the popular MSMT17~\cite{wei2018person} dataset using \cite{ge2020mutual} and set the output feature dimension to $256$.

\subsection{Individual framework validation} \label{subsec:individual}
We tried to use a person-tracking dataset, \textit{i.e.}, PersonPath22~\cite{personpath22}, to analyze the individual framework, however, we realized that this dataset (which has the same format as other visual person-tracking datasets~\cite{han2023mmptrack}) is not suitable for the evaluation of our framework for the following reasons:
\begin{itemize}
    \item The videos do not represent a human-robot collaboration scenario but are mainly composed of security cameras or person-view videos in crowded environments where people are far from the camera and appear in the image only for small instants of time.
    \item The videos are too short to correctly validate the re-identification module.
    \item In almost all the cases, the MOT algorithm we used in our framework already performed the right tracking because, when people disappear from the camera field-of-view (FOV), they do not enter it anymore, making unnecessary our re-identification module.
\end{itemize}
Due to such limitations, we decided to acquire a custom dataset for the validation. This dataset comprises $18$ videos for a total of $53$ minutes. Moreover, in the videos in which more actors are present, the framework is evaluated one time for each person, which brings the total minutes analyzed to $113$. The videos capture different working scenarios in two laboratory setups where a single person or group is present in the image. People in the videos are asked to exit and re-enter in camera FOV and to change their appearance (\textit{e.g.}, by wearing a sweatshirt) to test both the capabilities to re-identify when partial or total occlusions occur and adapt to new target appearances.
\begin{figure}
    \begin{subfigure}[b]{0.5\textwidth}
        \includegraphics[width=\linewidth]{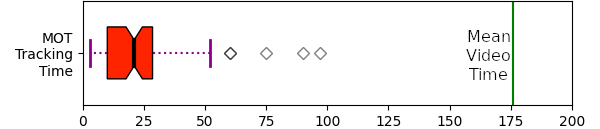}
        \caption{MOT tracking time (in seconds) w.r.t. mean video time.}
        \label{fig:time_plot_mot}
    \end{subfigure}\\
    \\
    \begin{subfigure}[b]{0.5\textwidth}
        \includegraphics[width=\linewidth]{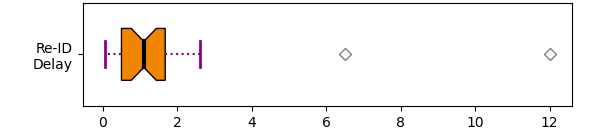}
        \caption{CARPE-ID delay (in seconds) before target re-identification.}
        \label{fig:time_plot_reid}
    \end{subfigure}\\
    \\
    \begin{subfigure}[b]{0.5\textwidth}
        \includegraphics[width=\linewidth]{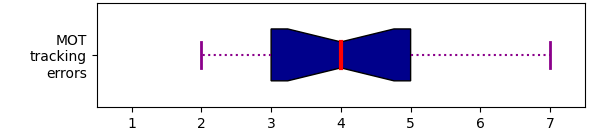}
        \caption{MOT number of errors for each video experiment.}
        \label{fig:error_plot_mot}
    \end{subfigure}\\
    \\
    \begin{subfigure}[b]{0.5\textwidth}
        \includegraphics[width=\linewidth]{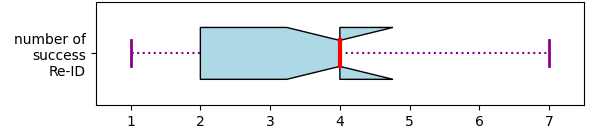}
        \caption{CARPE-ID number of re-identification for each video experiment.}
        \label{fig:error_plot_reid}
    \end{subfigure}
    \caption{Statistical evaluation of the obtained results.}
    \label{fig:plots}
\end{figure}
The analysis of these videos provided the following performances (more details are reported in Fig.~\ref{fig:plots}):
\begin{itemize}
     \item The min, mean, and max tracking lengths for the state-of-the-art MOT were $3.02$, $21.2$, and $52.2$ seconds (with some outliers which extended the max to $97.17$ seconds).
    \item The min, mean, and max Re-ID delay, \textit{i.e.} the time the framework takes to re-identify the target, was $0.06$, $1.1$, and $2.6$ seconds. There have been two cases in which the tracker took $6.5$ and $12.1$ seconds for the re-identification even if the target was in the scene.
    \item The MOT algorithm had min and max failure rates of $2$ and $7$ times with a mean failure rate of $4$ times for each video. Instead, our framework failed only $2$ times for \textit{all} the videos which demonstrate high robustness to partial and total occlusions.
    \item The min, mean, and max number of re-identification by our framework were $1$, $4$, and $7$.
\end{itemize}

\subsection{Application validation} \label{subsec:whole}

\begin{figure}
    \centering
    \includegraphics[width=\linewidth]{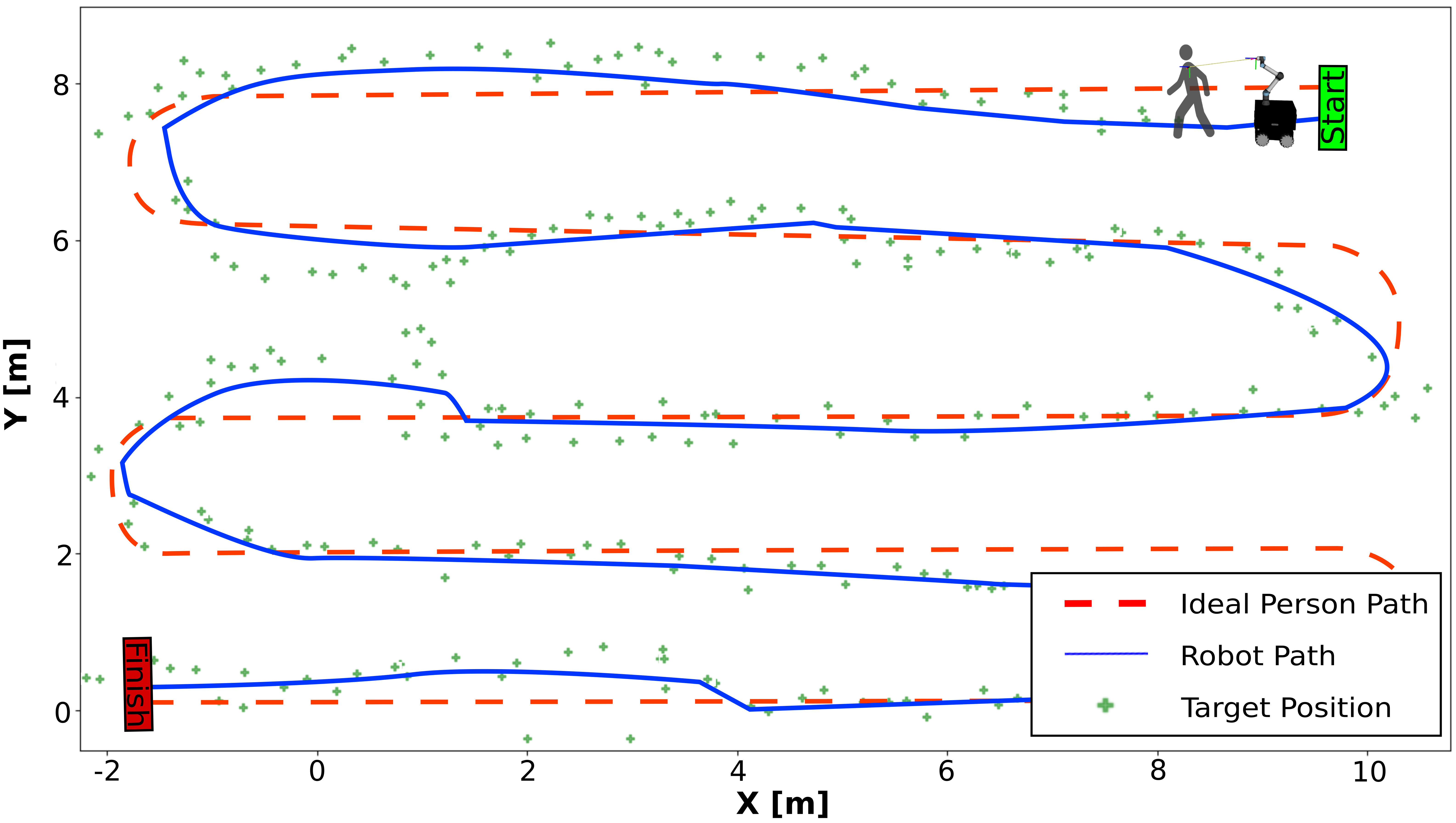}
    \caption{A FollowMe experiment sample: the target person has to roughly follow an ideal path (red dashed line) while the robot (blue line) has to follow him/her. The target positions computed using the CARPE-ID framework tracking are represented with green plus signs. The robot is initially placed in the green start position and has to follow the target until the red finish position is reached.}
    \label{fig:followme}
\end{figure}

To validate the framework on a real application we set up a robot person-following scenario in the same way as in \cite{rollo2023followme}. The target person is asked to move around and the robot has to follow him/her avoiding obstacles. However, in our case, the targets change their appearances during the following (condition not satisfied in \cite{rollo2023followme}) and pass among obstacles to test robustness to partial or total occlusions.
We stated that in \textit{all} the experiments the robot was able to correctly track, re-identify and follow the target person which moved the robot around in a laboratory setup where other people were working. In Fig.~\ref{fig:followme}, an experiment sample is plotted to give an idea of the experimentation setup. The target has to roughly follow the red dashed line, while the robot (blue line) has to follow him/her from start to finish. We performed the experiment $10$ times with $5$ different people as targets for a total of $837$ meters of following. A video example can be found \href{https://www.youtube.com/watch?v=xxsi8EkUcII&list=PLdibjJfM06ztpaKqJhrkF3fJxqXhRwL00&index=2&ab_channel=FedericoRollo}{here}.

\subsection{Results discussion} \label{subsect:res_discussion}
In this section, we analyze the results obtained from the considered evaluation tests to demonstrate the contributions presented in our work.
The individual results obtained in Sect.~\ref{subsec:individual}, clarify that human-robot interaction tasks require a personalized approach for re-identifying the target person. These application scenarios often require more specific recognition capabilities, \textit{e.g.}, when the person exits and re-enters in the camera FOV. 
For this reason, common SOT and MOT algorithms, are not enough for such scenarios and a more robust framework is necessary. 
This statement can be supported by the results presented in Fig.~\ref{fig:time_plot_mot} where we can see that the MOT algorithm has a mean tracking time of $21.2$ seconds while the mean video time is $176$ seconds and the minimum video time is $94$ seconds. 
In Fig.~\ref{fig:time_plot_reid}, we presented the delay between the target person's appearance in the camera FOV and the time in which the target is re-identified. The mean delay is $1.1$ seconds which is acceptable for an HRI application scenario. There have been $2$ cases in which the re-identifier was not able to re-identify the target in less than $6$ seconds. These cases happened due to a limitation of this framework: the target person should not change his/her appearance outside of the camera's field of view. In these cases, the targets re-entered into the camera's field of view with a different appearance \textit{e.g.}, they opened their sweatshirt, and the framework was not able to re-identify them with a non-negligible delay, therefore, they were considered as errors. Instead, in Fig.~\ref{fig:error_plot_mot} and Fig.~\ref{fig:error_plot_reid}, the times of the MOT algorithm losing track of the target are compared with the number of times the Re-ID framework was able to re-identify the target. The fact that both the plots have a mean of $4$ indicates that most of the MOT errors are recovered by our framework. Anyway, from the figure is clear that there are more MOT errors than Re-identifications. This is caused by the quick ID switching of the MOT algorithm and the re-identification delay; \textit{e.g.}, in a dataset video, the MOT changed the three target IDs in $1.4$ seconds and the re-identifier was able to re-identify only the last one in real-time.

The application scenario considers an example of a possible application for our framework. The video linked above and Fig.~\ref{fig:followme} show example experiments where the robot constantly follows the target in a laboratory environment where other people are working. Except for some small delays in the re-identification, the robot was always able to accomplish the following task for \textit{all} the experiments without human intervention.

\section{Conclusion and Future Works}
    \label{sec:conclusion}
    In this work, we presented a continuously adaptable person re-identification framework that can be used along with other human-robot interaction applications to enhance person-robot co-existence. We presented our pipeline composed of a MOT algorithm for detection and preliminary person tracking, a feature extraction network for appearance representation of the detected people and a re-identification structure based on a statistical distance, 
 an adaptable ideal target representation 
and threshold 
computation with a custom version of the exponential moving average where we added a damping factor (DEMA - damped exponential moving average). Using this setup we can perform person tracking when partial or total occlusions occur as well as when the target person changes his/her appearance during the tracking (\textit{e.g.}, by wearing a sweatshirt) which are challenging situations to be handed by MOT algorithms, as the experiments revealed.

A limitation of this work is that it suffers from catastrophic forgetting, \textit{i.e.}, the algorithm forgets the appearance of the target as time passes because it adapts to the new information it gathers. To overcome this problem a Continual Learning approach can be implemented where the feature extractor network can be trained online using a batch of target and non-target people images acquired during the re-identification. Such a solution could enhance the ability of the feature extractor to personalize the output features on the target's common appearance.
Another limitation is the difficulty of recognizing people at really high distances. This situation does not occur often in human-robot collaboration scenarios because the human is generally near to the robot and the camera but it is reasonable to highlight this point.

Future improvements will consider the enhancement of the re-identification capabilities when the target changes appearance outside the camera FOV. These cases could be recovered using a specialized network which can recognize the target through face identification.
Another improvement is to implement the continuous learning approach cited above to make the algorithm more robust. This technique can specialize the feature extraction network on the target appearance with online self-supervised training using the images of the target and of other people (the distractors).

\bibliographystyle{plain}
\bibliography{biblio}

\end{document}